\pdfoutput=1
\documentclass[11pt,a4paper]{article}
\usepackage[hyperref]{acl2019}
\usepackage{times}
\usepackage{latexsym}
\usepackage{multirow}
\usepackage{url}
\usepackage{color}
\usepackage{amsmath}
\usepackage{multirow}
\usepackage{graphicx}
\usepackage{verbatim}

\definecolor{mygreen}{RGB}{126,224,129}
\DeclareMathOperator{\argmax}{arg\,max}

\aclfinalcopy

\author{
  David Vilares\\
  Universidade da Coru\~{n}a, CITIC \\
  Departamento de Computaci\'{o}n \\
  Campus de Elvi\~{n}a s/n, 15071 \\ A Coru\~{n}a, Spain \\
  {\tt david.vilares@udc.es} \\
  \\\And
  Carlos G\'{o}mez-Rodr\'{i}guez \\
  Universidade da Coru\~{n}a, CITIC \\
  Departamento de Computaci\'{o}n \\
  Campus de Elvi\~{n}a s/n, 15071 \\ A Coru\~{n}a, Spain \\
  {\tt carlos.gomez@udc.es}  \\}

\begin{document}

\title{HEAD-QA: A Healthcare Dataset for Complex Reasoning}

\maketitle
\begin{abstract}

We present \textsc{head-qa}, a multi-choice question answering testbed to encourage research on complex reasoning. The questions come from exams to access a specialized position in the Spanish healthcare system, and are challenging even for highly specialized humans.
We then consider monolingual (Spanish) and cross-lingual (to English) experiments with information retrieval and neural techniques. We show that: 
(i) \textsc{head-qa} challenges current methods, 
and (ii) the results lag well behind human performance, demonstrating its usefulness as a benchmark for future work.

\end{abstract}

\section{Introduction}

Recent progress in question answering (\textsc{qa}) has been led by neural models \cite{seo2016bidirectional,kundu2018nil}, due to their ability to process raw texts. 
However, some authors \cite{kaushik2018much,clark2018think} have discussed  the tendency of research to develop datasets and methods that accomodate the data-intensiveness and strengths of \emph{current} neural methods. 

This is the case of popular English datasets such as bAbI \cite{weston2015towards} or SQuAD \cite{rajpurkar2016squad,rajpurkar2018know}, where some systems achieve near human-level performance \cite{hu2017reinforced,xiong2017dcn+} and often surface-level knowledge suffices to  answer.
To counteract this, \newcite{clark2016combining} and \newcite{clark2018think} have encouraged progress by developing multi-choice datasets that require reasoning. The questions match grade-school science, due to the difficulties to collect specialized questions. 
With a similar aim, \newcite{lai2017race} released 100k questions and 28k passages intended for middle or high school Chinese students, and \newcite{zellers2018swag} introduced a dataset for common sense reasoning from a spectrum of daily situations.

However, this kind of dataset is scarce for complex domains like medicine: while challenges have been proposed in such domains, like textual entailment \cite{abacha2015semantic,abacha2016recognizing} or answering questions about specific documents and snippets \cite{bioasq2018}, we know of no resources that require general reasoning on complex domains.
The novelty of this work falls in this direction, presenting a multi-choice \textsc{qa} task that combines the need of knowledge and reasoning with complex domains, and which takes humans years of training to answer correctly.

\begin{table}
\small{
\centering
\begin{tabular}{p{7.3cm}}
\textbf{Question (medicine)}: A 13-year-old girl is operated on due to Hirschsprung illness at 3 months of age. Which of the following tumors is more likely to be present?\\ 

1. Abdominal neuroblastoma\\
2. \colorbox{mygreen}{Wilms tumor}\\
3. Mesoblastic nephroma\\
4. Familial thyroid medullary carcinoma.\\
\\

\textbf{Question (pharmacology)} The antibiotic treatment of choice for Meningitis caused by Haemophilus influenzae serogroup b is:\\
1. Gentamicin \\
2. Erythromycin \\
3. Ciprofloxacin \\
4. \colorbox{mygreen}{Cefotaxime} \\
\\
\textbf{Question (psychology)}	According to research derived from the Eysenck model, there is evidence that extraverts, in comparison with introverts:\\
1. Perform better in surveillance tasks.\\
2. Have greater salivary secretion before the lemon juice test.\\
3. \colorbox{mygreen}{Have a greater need for stimulation.}\\
4. Have less tolerance to pain.\\

\end{tabular}}
\caption{Samples from \textsc{head-qa}}\label{table-sample-head-qa}

\end{table}

\paragraph{Contribution} 
We present \textsc{head-qa}, a multi-choice testbed of graduate-level questions about medicine, nursing, biology, chemistry, psychology, and pharmacology (see Table \ref{table-sample-head-qa}\footnote{These examples were translated by humans to English.}). The data is in Spanish, but we also include an English version.
We then test models for open-domain and multi-choice \textsc{qa}, showing the complexity of the dataset and its utility to encourage progress in \textsc{qa}. \textsc{head-qa} and models can be found at \url{http://aghie.github.io/head-qa/}.

\section{The \textsc{head-qa} corpus}\label{section-corpus}

The Ministerio de Sanidad, Consumo y Bienestar Social\footnote{\url{https://www.mscbs.gob.es/}} (as a part of the Spanish government) announces every year examinations to apply for specialization positions in its public healthcare areas.
The applicants must have a bachelor's degree in the corresponding area (from 4 to 6 years) and they prepare the exam for a period of one year or more, as the vacancies are limited. The exams are used to discriminate among thousands of applicants, who will choose a specialization and location according to their mark (e.g., in medicine, to access a cardiology or gynecology position at a given hospital).

\begin{table}
\tabcolsep=0.15cm
\centering
\small{
\begin{tabular}{lcccc}
\hline
\bf \multirow{2}{*}{Category} & \bf Unsupervised & \multicolumn{3}{c}{\bf Supervised setting} \\
&\bf setting & \bf Train & \bf Dev & \bf Test \\
\hline
Biology & 1,132 & 452 & 226 & 454\\
nursing & 1,069 & 384 & 230 & 455\\
Pharmacology & 1,139 & 457 &225 & 457\\
Medicine & 1149  & 455 & 231 & 463\\
Psychology & 1134 & 453 & 226 & 455\\
Chemistry & 1142 & 456 & 228 & 458\\
\hline
Total & 6,765 & 2,657 & 1,366 & 2,742 \\
\hline

\end{tabular}}
\caption{Number of questions in \textsc{head-qa}}\label{table-head-stats}
\end{table}

\begin{table}
\tabcolsep=0.14cm
\centering
\small{
\begin{tabular}{lcccc}
\hline
\bf \multirow{2}{*}{Category} & \bf Longest & \bf Avg & \bf Longest &\bf Avg \\
&\bf question &\bf question&\bf answer & \bf answer\\
\hline
Biology & 43 & 11.11 & 40 & 5.08 \\
Nursing &187 & 29.03& 94 & 9.54 \\
Pharmacology & 104 & 18.18 & 43 & 6.70 \\
Medicine &308 & 55.29 & 85 & 9.31 \\
Psychology & 103 & 21.91 & 43 & 7.98 \\
Chemistry & 63 & 15.82 & 52 & 7.62 \\
\hline
\end{tabular}}
\caption{Tokens statistics in \textsc{head-qa}}\label{table-head-stats-2}
\end{table}
 
\begin{figure}[tbp]
    \centering
    \includegraphics[width=0.5\columnwidth]{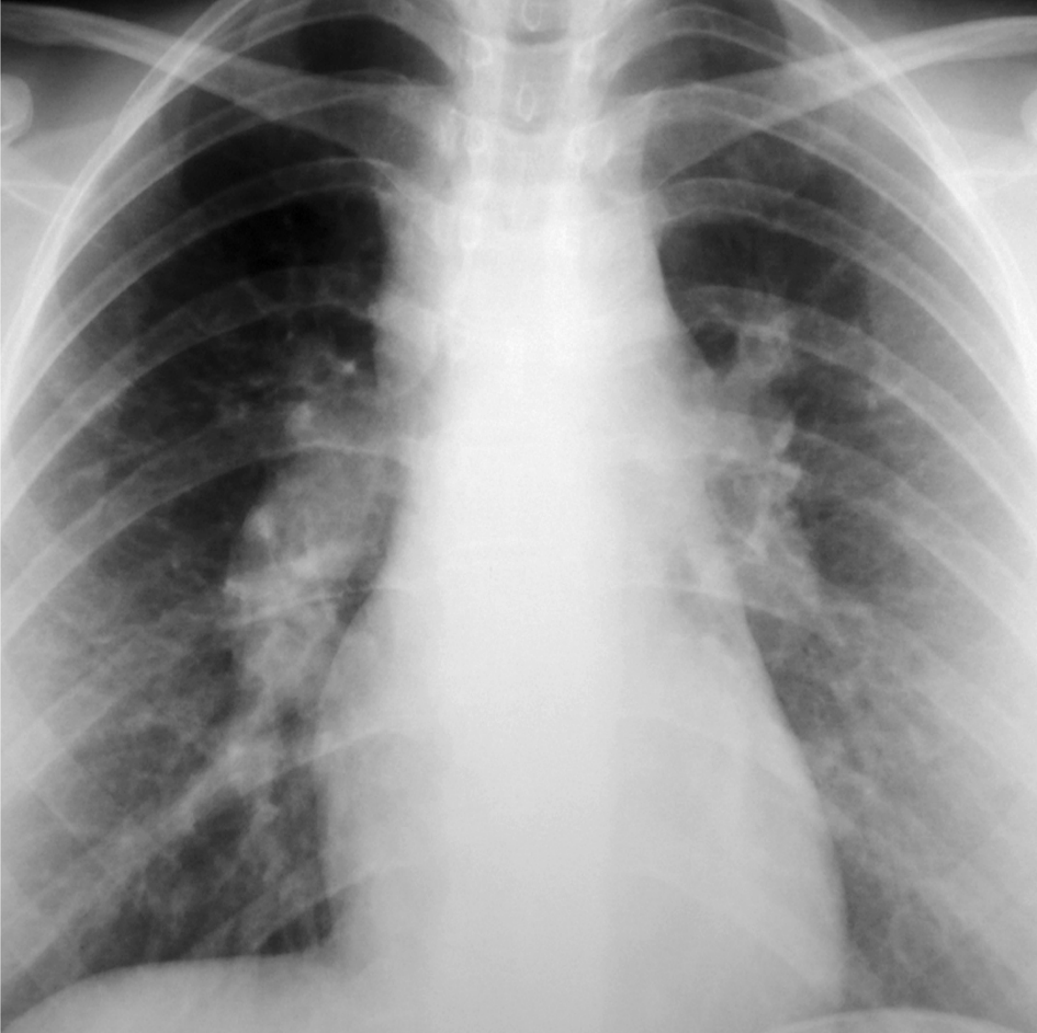}
    \caption{Image no 21 from MIR 2017}\label{fig-21-mir-2017}

\end{figure}

We use these examinations (from 2013 to present) to create \textsc{head-qa}. We consider questions involving the following healthcare areas: medicine (aka \textsc{mir}), pharmacology (\textsc{fir}), psychology (\textsc{pir}), nursing (\textsc{eir}), biology (\textsc{bir}), and chemistry (\textsc{qir}).\footnote{Radiophysics exams are excluded, due to the difficulty to parse their content (e.g. equations) from the \textsc{pdf} files.}\footnote{Some of the questions might be considered invalid after the exams. We remove those questions from the final dataset.} 
 Exams from 2013 and 2014 are multi-choice tests with five options, while the rest of them have just four. The questions mainly refer to technical matters, although some of them also consider social issues (e.g. how to deal with patients in stressful situations). A small percentage ($\sim$14\%) of the medicine questions refer to images that provide additional information to answer correctly. These are included as a part of the corpus, although we will not exploit them in this work. For clarity, Table \ref{table-question-image} shows an example:\footnote{Note that images often correspond to serious injuries and diseases.
 Viewer discretion is advised. The quality of the images varies widely, but it is good enough that the pictures can be analyzed by humans in a printed version. Figure \ref{fig-21-mir-2017} has 1037x1033 pixels.}

  \begin{minipage}[b]{\columnwidth}
    \centering
    \begin{tabular}{p{6.6cm}}
    \vspace{0.2cm}
    \small{\textbf{Question} Question linked to image no 21. A 38-year-old bank employee who has been periodically checked by her company is referred to us to assess the chest X-ray. The patient smokes 20 cigarettes / day from the age of 21. She says that during the last months, she is somewhat more tired than usual. The basic laboratory tests are normal except for an Hb of 11.4 g / dL. An electrocardiogram and forced spirometry are normal. What do you think is the most plausible diagnostic orientation?}\\ 
    \vspace{0.2cm}
    \small{1. Hodgkin's disease.}\\
    \small{2. Histoplasmosis type fungal infection.}\\
    \small{3. \colorbox{mygreen}{Sarcoidosis.}}\\
    \small{4. Bronchogenic carcinoma.}\\
    \end{tabular}
    \captionof{table}{A question referring to Figure \ref{fig-21-mir-2017}}\label{table-question-image}
    \end{minipage}
    
We describe in detail the \textsc{json} structure of \textsc{head-qa} in Appendix \ref{section-appendix}. We enumerate below the fields for a given sample:
\begin{itemize}
    \item The question \textsc{id} and the question's content.
    \item Path to the image referred to in the question (if any).
    \item A list with the possible answers. Each answer is composed of the answer ID and its text.
    \item The \textsc{id} of the right answer for that question.
\end{itemize}

Although all the approaches that we will be testing are unsupervised or distant-supervised, we additionally define official training, development and test splits, so future research with supervised approaches can be compared with the work presented here. For this supervised setting, we choose the 2013 and 2014 exams for the training set, 2015 for the development set, and the rest for testing. The statistics are shown in Tables \ref{table-head-stats} and \ref{table-head-stats-2}. It is worth noting that a common practice to divide a dataset is to rely on randomized splits to avoid potential biases in the collected data. We decided not to follow this strategy for two reasons. First, the questions and the number of questions per area are designed by a team of healthcare experts who already try to avoid these biases. Second (and more relevant), random splits would impede comparison against official (and aggregated) human results.

Finally, we hope to increase the size of \textsc{head-qa} by including questions from future exams.

\paragraph{English version} \textsc{head-qa} is in Spanish, but we include a translation to English (\textsc{head-qa-en}) using the Google API, which we use to perform cross-lingual experiments. We evaluated the quality of the translation using a sample of 60 random questions and their answers. We relied on two fluent Spanish-English speakers to score the adequacy\footnote{Adequacy: How much meaning is preserved? We use a scale from 5 to 1: 5 (all meaning), 4 (most meaning), 3 (some meaning), 2 (little meaning), 1 (none).} and on one native English speaker for the fluency,\footnote{Fluency: Is the language in the output fluent? We use a scale from 5 to 1: 5 (flawless), 4 (good), 3 (non-native), 2 (disfluent), 1 (incomprehensible).} following the scale by \newcite{koehn2006manual}. The average scores for adequacy were 4.35 and 4.71 out of 5, i.e. most of the meaning is captured; and for fluency 4 out of 5, i.e. good. As a side note, it was observed by the annotators that 
most
names of diseases were successfully translated to English. On the negative side, the translator tended to struggle with elements such as molecular formulae, relatively common in chemistry questions.\footnote{This particular issue is not only due to the automatic translation process, but also to the difficulty of correctly mapping these elements from PDF exams to plain text.}

\section{Methods}\label{section-open-domain-question-answering solvers}

\paragraph{Notation} We represent \textsc{head-qa} as a list of tuples: $[(q_0,A_0), ..., (q_N,A_N)]$, where:  $q_i$ is a question and $A_i = [a_{i0},...,a_{im}]$ are the possible answers. We use $\tilde{a}_{ik}$ to denote the predicted answer, ignoring indexes when not needed.\\

\noindent\newcite{kaushik2018much} discuss on the need of providing rigorous baselines that help better understand the improvement coming from future models, and also the need of avoiding architectural novelty when introducing new datasets. For this reason, our baselines are based on state-of-the-art systems used in open-domain and multi-choice \textsc{qa} \cite{chen2017reading,kembhavi2017you,khot2018scitail,clark2018think}.

\subsection{Control methods}

Given the complex nature of the task, we include three control methods:

\paragraph{Random} Sampling $\tilde{a} \sim \mathit{Multinomial}(\phi)$, where $\phi$ is a random distribution.
\paragraph{Blind$_x$} $\tilde{a}_{ik}$ = $a_{ix}\ \forall i$. Always chosing the $x$th option. Tests made by the examiners are not totally random \cite{poundstone2014rock} and right answers tend occur more in middle options.
\paragraph{Length} Choosing the longest answer.\footnote{Computed as the number of characters in the answers.} \newcite{poundstone2014rock} points out that examiners have to make sure that the right answer is totally correct, which might take more space.

\subsection{Strong multi-choice methods}

We evaluate an information retrieval (\textsc{ir}) model for \textsc{head-qa} and cross-lingual models for \textsc{head-qa-en}. Following \newcite{chen2017reading}, we use Wikipedia as our source of information ($\mathcal{D}$)\footnote{We downloaded Spanish and English Wikipedia dumps.} for all the baselines. We then extract the raw text and remove the elements that add some type of structure (headers, tables, \dots).\footnote{ \url{github.com/attardi/wikiextractor}} 

\subsubsection{Spanish information retrieval}

Let $(q_i, [a_{i0},...,a_{im}])$ be a question with its possible answers, we first create a set of $m$ queries of the form $[q_i+a_{i0}, ..., q_i+a_{im}]$, which will be sent separately to a search engine. In particular, we use the DrQA's Document Retriever \cite{chen2017reading}, which scores the relation between the queries and the articles as \textsc{tf-idf} weighted bag-of-word vectors, and also takes into account word order and bi-gram counting. The predicted answer is defined as $\tilde{a}_{ik}$ = $\argmax_k(score(m_k, \mathcal{D}))$, i.e. the answer in the query $m_k$ for which we obtained the highest document relevance. This is equivalent to the \textsc{ir} baselines by \newcite{clark2016combining,clark2018think}.

\subsubsection{Cross-lingual methods}

Although some research on Spanish \textsc{qa} has been done in the last decade
\cite{magnini2003multiple,vicedo2003question,buscaldi2006mining,kamateri2019test}, most recent work has been done for English, in part due to the larger availability of resources. On the one hand this is interesting because we hope \textsc{head-qa} will encourage research on multilingual question answering. On the other hand, we want to check how challenging the dataset is for state-of-the-art systems, usually available only for English. To do so, we use \textsc{head-qa-en}, as the adequacy and the fluency scores of the translation were high.

\paragraph{Cross-lingual Information Retrieval} The \textsc{ir} baseline, but applied to \textsc{head-qa-en}. We also use this baseline as an extrinsic way to evaluate the quality of the translation, expecting to obtain a performance similar to the Spanish \textsc{ir} model.

\paragraph{Multi-choice DrQA} \cite{chen2017reading} DrQA first returns the 5 most relevant documents for each question, relying on the information retrieval system described above. It will then try to find the exact span in them containing the right answer on such documents, using a document reader. For this, the authors rely on a neural network system inspired in the Attentive Reader \cite{hermann2015teaching} that was trained over SQuAD \cite{rajpurkar2016squad}. The original DrQA is intended for open-domain \textsc{qa}, focusing on factoid questions. To adapt it to a multi-choice setup, to select $\tilde{a}$ we compare the selected span against all the answers and select the one that shares the largest percentage of tokens.\footnote{We lemmatize and remove the stopwords as in \cite{clark2018think}. We however observed that many of selected spans did not have any word in common with any of the answers. If this happens, we select the longest answer.} Non-factoid questions (common in \textsc{head-qa}) are not given any special treatment.

\paragraph{Multi-choice BiDAF} \cite{clark2018think} Similar to the multi-choice DrQA, but using a BiDAF architecture as the document reader \cite{seo2016bidirectional}. The way BiDAF is trained is also different: they first trained the reader on SQuAD, but then further tuned to science questions presented in \cite{clark2018think}, using continued training. This system might select as correct more than one answer. If this happens, we follow a simple approach and select the longest one.

\paragraph{Multi-choice DGEM and Decompatt} \cite{clark2018think} The models adapt the DGEM \cite{decompatt} and Decompatt \cite{khot2018scitail} entailment systems. They consider a set of hypothesis $h_{ik}$=$q_i + a_{ik}$ and each $h_i$ is used as a query to retrieve a set of relevant sentences, $\mathcal{S}_{ik}$. Then, an entailment score $\mathit{entailmen}t(h_{ik},s)$ is computed for every $h_{ik}$ and $s \in S_{ik}$, where $\tilde{a}$ is the answer inside $h_{ik}$ that maximizes the score. If multiple answers are selected, we choose the longest one.

\section{Experiments}\label{section-experiments}

\paragraph{Metrics} We use accuracy and a \textsc{points} metric (used in the official exams): a right answer counts 3 points and a wrong one subtracts 1 point.\footnote{Note that as some exams have more choices than others, there is not a direct correspondence between accuracy and \textsc{points} (a given healthcare area might have better accuracy than another one, but worse \textsc{points} score).}

\paragraph{Results (unsupervised setting)} Tables \ref{table-ES-accuracy} and \ref{table-ES-points} show the accuracy and \textsc{points} scores for both \textsc{head-qa} and \textsc{head-qa-en}. The cross-lingual \textsc{ir} model obtains even a greater performance than the Spanish one. This is another indicator that the translation is good enough to apply cross-lingual approaches. On the negative side, the approaches based on current neural architectures obtain a lower performance. 

\begin{table}[hbtp]
\tabcolsep=0.095cm
\small{
\centering
\begin{tabular}{l|l|ccccccc}
\hline
\bf & \bf Model & \bf \textsc{bir} & \bf \textsc{mir} & \bf \textsc{eir} & \bf \textsc{fir} & \bf \textsc{pir} & \bf \textsc{qir} & \bf Avg \\
\hline
\multirow{7}{*}{ES}&\textsc{random}   & 24.2 & 22.0 & 25.1 & 23.2 & 24.0 & 24.5 & 23.8 \\
&\textsc{blind$_1$} & 23.7 & 22.8 & 22.7 & 22.4 & 22.5 & 21.2 & 22.5 \\
&\textsc{blind$_2$} & 25.6 & 24.3 & 23.5 & 23.0 & 25.3 & 24.9 & 24.4 \\
&\textsc{blind$_3$} & 23.0 & 24.7 & 26.5 & 25.8 & 22.9 & 25.1 & 24.7 \\
&\textsc{blind$_4$} & 22.6 & 20.0 & 21.7 & 22.4 & 23.2 & 22.5 & 22.1 \\
&\textsc{length}&26.9&24.9&28.6&28.7&30.6&29.0&28.1\\
&\textsc{ir}  & 34.5 & 26.5 & \bf 32.7 & 35.5 & 34.2 & \bf 34.2 & 32.9 \\
\hline
\multirow{5}{*}{EN}&\textsc{ir} & \bf 37.9 & \bf 30.3 & 32.6 & \bf 38.7 & \bf 34.7 & 33.7 & \bf 34.6 \\
&\textsc{drqa} & 29.5 & 25.0 & 27.3 & 28.3 & 31.0 & 30.2 & 28.5 \\
&\textsc{bidaf} & 33.4 & 26.2 & 26.8 & 29.9 & 26.8 & 30.3 & 28.9 \\
&\textsc{dgem}  & 31.7 & 25.7 & 28.7 & 29.9 & 28.5 & 30.3 & 29.1 \\
&\textsc{decompatt} & 30.6 & 23.6 & 27.9 & 27.2 & 28.3 & 27.6 & 27.5 \\
\hline

\end{tabular}}
\caption{Accuracy on the \textsc{head-qa} and \textsc{head-qa-en} corpora (unsupervised setting)}\label{table-ES-accuracy}
\end{table}

\begin{table}[hbtp]
\tabcolsep=0.06cm
\small{
\centering
\begin{tabular}{l|l|ccccccc}
\hline
&\bf Model & \bf \textsc{bir} & \bf \textsc{mir} & \bf \textsc{eir} & \bf \textsc{fir} & \bf \textsc{pir} & \bf \textsc{qir} & \bf Avg \\
\hline
\multirow{3}{*}{ES}&\textsc{blind$_3$} & -17.6 & -2.6 & 16.6& 7.4& -18.8 & 1.2 & -2.3\\
&\textsc{length}&16.8&-1.0&32.6&33.8&50.8&36.4&28.2\\
&\textsc{ir}  &  86.4 &  14.2 &  67.0 &  95.4 &  82.8 & \bf 84.4 &  71.7 \\
\hline
\multirow{5}{*}{EN}&\textsc{ir}  & \bf 116.8 & \bf 48.6 & \bf 67.8 & \bf 125.0  & \bf 87.6 & 79.6 & \bf 87.6  \\
&\textsc{drqa} & 40.8 & -0.2 & 20.6 & 29.8 & 54.0 & 47.6 & 32.1 \\
&\textsc{bidaf} & 75.6 & 11.0 & 15.8 & 44.4 & 16.6 & 48.6 & 35.3  \\
&\textsc{dgem}  & 60.8 & 7.0 & 34.2 & 45.0 & 31.6 & 48.4 & 37.8 \\
&\textsc{decompatt} & 51.2 & -13.0 & 27.8 & 20.2 & 30.0 & 23.6 & 23.3 \\
\hline

\end{tabular}}
\caption{\textsc{points} on the \textsc{head-qa} and \textsc{head-qa-en} corpora (unsupervised setting)}\label{table-ES-points}
\end{table}

\paragraph{Results (supervised setting)}
We show in Tables \ref{table-accuracy-supervised} and \ref{table-points-supervised} the performance of the top models on the test split corresponding to the supervised setting.

\begin{table}[hbtp]
\tabcolsep=0.12cm
\small{
\centering
\begin{tabular}{l|l|ccccccc}
\hline
&\bf Model & \bf \textsc{bir} & \bf \textsc{mir} & \bf \textsc{eir} & \bf \textsc{fir} & \bf \textsc{pir} & \bf \textsc{qir} & \bf Avg \\
\hline
\multirow{4}{*}{ES}&\textsc{random} & 24.2&23.1&25.2&23.8&27.9&27.7&25.3\\
&\textsc{blind}$_3$ &26.0&27.5&29.8&27.2&24.8&27.8&27.2\\
&\textsc{length} &32.4&27.0&32.8&30.2&30.5&30.1&30.5\\
&\textsc{ir} &36.5&26.3&36.0&40.3&\bf 35.9&\bf 36.2&35.2\\
\hline
\multirow{3}{*}{EN}&\textsc{ir}  &\bf 39.8&\bf 33.3&\bf 36.4&\bf 42.2&35.7&36.0&\bf 37.2\\
&\textsc{bidaf} & 36.5&26.6&27.7&29.3&28.1&34.1&30.3\\
&\textsc{dgem} &31.7&27.2&30.7&29.9&31.0&33.2&30.6\\
\hline

\end{tabular}}
\caption{Accuracy on the \textsc{head-qa} and \textsc{head-qa-en} corpora (supervised setting)}\label{table-accuracy-supervised}
\end{table}

\begin{table}[hbtp]
\tabcolsep=0.07cm
\small{
\centering
\begin{tabular}{l|l|ccccccc}
\hline
&\bf Model & \bf \textsc{bir} & \bf \textsc{mir} & \bf \textsc{eir} & \bf \textsc{fir} & \bf \textsc{pir} & \bf \textsc{qir} & \bf Avg \\
\hline
\multirow{4}{*}{ES}&\textsc{random} &-7.0&-17.5&2.5&-10-5&26.5&25.0&3.2\\
&\textsc{blind$_3$} &9.0&22.5&44.5&19.5&-1.5&25.0&19.8\\
&\textsc{length} &67.0&18.5&70.5&47.5&50.5&47.0&50.2\\
&\textsc{ir}  &105.0&12.5&100.5&139.5&\bf 98.5&\bf 103.0&93.2\\
\hline
\multirow{3}{*}{EN}&\textsc{ir}  & \bf 135.0& \bf 76.5& \bf 104.5& \bf 157.5& 96.5& 101.0& \bf 111.8\\
&\textsc{bidaf} & 104.0 & 14.5 & 18.5&39.0&29.0&83.0& 48.0\\
&\textsc{dgem} & 61.0 & 20.5 & 52.5 & 45.5 & 54.5 & 75.0 & 51.5\\
\hline

\end{tabular}}
\caption{\textsc{points} on the \textsc{head-qa} and \textsc{head-qa-en} corpora (supervised setting)}\label{table-points-supervised}
\end{table}

\paragraph{Discussion}

Medicine questions (\textsc{mir}) are the hardest ones to answer across the board. We believe this is due to the greater length of both the questions and the answers (this was shown in Table \ref {table-head-stats-2}). This hypothesis is supported by the lower results on the nursing  domain (\textsc{eir}), the category with the second longest questions/answers. On the contrary, the categories for which we obtained the better results, such as pharmacology (\textsc{fir}) or biology (\textsc{bir}), have shorter questions and answers. While the evaluated models surpass all control methods, their performance is still well behind the human performance. We illustrate this in Table \ref{table-human-performance}, comparing the performance (\textsc{points} score) of our best model against a summary of the results, on the 2016 exams.\footnote{2016 was the annual examination for which we were able to find more available information.} Also, the best performing model was a non-machine learning model based on standard information retrieval techniques.
This reinforces the need for effective information extraction techniques that can be later used to perform complex reasoning with machine learning models.

\begin{table}[hbtp]
\tabcolsep=0.13cm
\small{
\centering
\begin{tabular}{l|cccccc}
\hline
 & \bf \textsc{bir} & \bf \textsc{mir} & \bf \textsc{eir} & \bf \textsc{fir} & \bf \textsc{pir} & \bf \textsc{qir} \\
 \hline
Avg 10 best &\multirow{2}{*}{627.1} & \multirow{2}{*}{592.2} & \multirow{2}{*}{515.2} & \multirow{2}{*}{575.5} & \multirow{2}{*}{602.1} & \multirow{2}{*}{529.1} \\
humans & \\
Pass mark &  219.0 & 207.0 & 180.0 & 201.0 & 210.0 & 185.0 \\
\hline
EN \textsc{ir} & 168.0 & 124.0 & 77.0 & 132.0 & 62.0 & 93.0 \\
\hline

\end{tabular}}
\caption{Human performance on the 2016 exams. The results are not strictly comparable, as the last 10 questions are considered as backup questions in the human exams, but still show how far the tested baselines are from human performance.}\label{table-human-performance}
\end{table}

\section{Conclusion}

We presented a complex multi-choice dataset containing questions about medicine, nursing, biology, pharmacology, psychology and chemistry. Such questions correspond to examinations to access specialized positions in the Spanish healthcare system, and require specialized knowledge and reasoning to be answered. To check its complexity, we then tested different state-of-the-art models for open-domain and multi-choice questions. We show how they struggle with the challenge, being clearly surpassed by a non-machine learning model based on information retrieval. We hope this work will encourage research on designing more powerful \textsc{qa} systems that can carry out effective information extraction and reasoning. 

We also believe there is room for alternative challenges in \textsc{head-qa}. In this work we have used it as a \emph{closed} \textsc{qa} dataset (the potential answers are used as input to determine the right one). Nothing prevents to use the dataset in an \emph{open} setting, where the system is given no clue about the possible answers. This would require to think as well whether widely used metrics such as \textsc{bleu} \cite{papineni2002bleu} or exact match could be appropriate for this particular problem.

\section*{Acknowlegments}

This work has received support from the TELEPARES-UDC project
(FFI2014-51978-C2-2-R) and the ANSWER-ASAP project (TIN2017-85160-C2-1-R) from MINECO, from Xunta de Galicia (ED431B 2017/01), and from the European Research Council (ERC), under the European Union's Horizon 2020 research and innovation programme (FASTPARSE, grant agreement No 714150). We thank Mark Anderson for his help with translation fluency evaluation.

\bibliography{qa_health}
\bibliographystyle{acl_natbib}

\appendix

\section{Appendices}\label{section-appendix}

We below describe the \texttt{fields} of the \textsc{json} file use to represent \textsc{head-qa}.

\small{
\begin{verbatim}
{    
 "version": 1.0
 "language": ["es","en"]
 "exams": A list of exams.
    "name": Cuaderno_YEAR_1_*IR_ACRONYM.
    "year": e.g. 2016.
    "category": ['medicine','biology',
                 'nursing','pharmacology',
                 'chemistry','psychology'] 
    "data": A list of questions/answers.
       "qid": The question ID, extracted 
              from the original PDF exam 
              (usually between 1 and 235).
       "qtext" : The text of the question.
       "ra" : The ID of the right answer.
       "answers": A list with the answer options.
          "aid": The answer ID (1 to 5).
          "atext": The text of the answer.
}
\end{verbatim}
\label{listing-json-head-qa}
}
\end{document}